\documentclass[11pt]{article}

\usepackage[utf8]{inputenc}
\usepackage[T1]{fontenc}
\usepackage{amsmath,amssymb,amsfonts}
\usepackage{graphicx}
\usepackage{booktabs}
\usepackage{multirow}
\usepackage{hyperref}
\usepackage{xcolor}
\usepackage{microtype}
\usepackage{geometry}
\usepackage{natbib}
\usepackage{caption}
\usepackage{subcaption}
\usepackage{enumitem}
\usepackage{array}
\usepackage{tabularx}

\title{Recover-LoRA for Aggressive Quantization: Reclaiming Accuracy\\in 2-Bit Language Models via Low-Rank Adaptation\\with Knowledge Distillation on Synthetic Data}

\author{
  Devleena Das \quad Rajeev Patwari \quad Elliott Delaye \quad Ashish Sirasao \\
  Advanced Micro Devices, Inc.\ (AMD) \\
  \texttt{\{devleena.das, rajeev.patwari, elliott.delaye, ashish.sirasao\}@amd.com}
}

\date{}

\begin{document}
\maketitle

\begin{abstract}
Aggressive weight quantization to 2-bit precision offers substantial throughput and memory gains for large language model (LLM) inference, but typically incurs severe accuracy degradation.
These gains are particularly relevant for edge and on-device deployment, where memory capacity and bandwidth are primary constraints.
In this work, we extend Recover-LoRA---a lightweight, data-free accuracy recovery method originally developed for general model weight corruption---to the setting of ultra-low-bit quantization.
We propose a selective mixed-precision strategy in which only gate and up projection layers of the MLP are quantized to 2-bit (W2), while all other linear layers remain at higher precision, yielding a mixed-precision GateUp configuration.
We demonstrate via roofline analysis across three model families (4B--20B) and two hardware platforms that a W4/W2-GateUp deployment (4-bit base with 2-bit gate/up) delivers 7.5--23.3\% TPS improvement over uniform W4 depending on model and context length, while confining quantization error to a predictable subset of layers.
We then apply Recover-LoRA---training low-rank adapters on the quantized layers via logit distillation with synthetic data---to recover accuracy lost from 2-bit quantization of the gate and up layers.
In a case study on Qwen3-4B, Recover-LoRA achieves 80--95\% accuracy recovery on 9 of 12 benchmarks, using only 10k synthetic training samples and no labeled data.
We further demonstrate that synthetic data performs comparably to curated labeled data for distillation-based recovery, and that recovery generalizes to out-of-distribution evaluation tasks.
Our results present Recover-LoRA as a practical post-quantization accuracy recovery tool for aggressive weight compression in deployment settings.
\end{abstract}

\section{Introduction}

Large language models (LLMs) continue to scale rapidly, with recent models such as Qwen3-235B~\citep{yang2025qwen3}, DeepSeek-V3~\citep{deepseekv3}, and Llama~3~\citep{grattafiori2024llama3} reaching hundreds of billions of parameters.
At these scales, aggressive compression is necessary to reduce serving costs and improve throughput in data center environments.
At smaller scales, 1B to 20B parameters, compression plays a different but equally important role, enabling deployment on edge and on-device platforms where memory capacity and bandwidth are hard constraints.
In both regimes, weight quantization remains a primary compression technique.
Methods such as AWQ~\citep{lin2024awq} and GPTQ~\citep{frantar2023gptq} have established 4-bit quantization as near-lossless for many model families, and recent work has begun exploring sub-4-bit precision, such as QuIP\#~\citep{tseng2024quip} and AQLM~\citep{egiazarian2024aqlm} at 2--3 bits, to further reduce memory footprint and improve inference throughput.
However, reducing to 2-bit precision remains challenging in practice since the increased quantization noise leads to significant accuracy degradation, limiting the practical adoption of ultra-low-bit models~\citep{zhu2024survey}.
At the same time, the throughput benefits of lower-precision weights are well-established. In the decode-bound regime, inference throughput is primarily limited by memory bandwidth for loading model weights~\citep{patwari2025life}, and reducing weight precision from 4-bit to 2-bit directly reduces the bytes transferred per token, yielding meaningful TPS improvements. For practical deployment, this raises a critical question: \emph{if 2-bit quantization delivers meaningful throughput and memory gains but degrades accuracy, can that accuracy be recovered cheaply, without full model retraining or access to labeled data, for widespread usability?}

In this work, we address the accuracy recovery challenge by framing it as a two-stage problem: (1) selective post-training quantization (PTQ) followed by (2) lightweight post-hoc recovery using low-rank adaptation and knowledge distillation via Recover-LoRA~\citep{das2025recoverlora}.
We first present a mixed-precision strategy that exploits the uneven contribution of transformer layers to throughput.
The gate and up projection layers in the MLP blocks account for the majority of model parameters in modern gated-MLP architectures and are a primary throughput bottleneck during decode.
By selectively quantizing only these layers to 2-bit while keeping remaining layers at 4-bit, a configuration we term \textbf{W4/W2-GateUp}, we capture a substantial portion of the throughput benefit of full W2 quantization (7.5--23.3\% TPS gains over W4 across the three model families studied, depending on model and context length).
Similarly, recent component-level sensitivity analyses confirm that MLP projections are the most quantization-sensitive components in transformer architectures, while attention projections are substantially more robust~\citep{cim2026diagnosing}.
Targeting gate and up layers for aggressive quantization therefore maximizes throughput gains but also incurs meaningful accuracy degradation, which motivates the second stage of our approach.
In stage two, we apply \textbf{Recover-LoRA} to recover the lost accuracy.
Since the quantized layers are known a priori, we attach low-rank adapters (LoRA) directly to the gate and up projections and train them via knowledge distillation against the full-precision model, using only synthetically generated data, requiring no labeled datasets and no full model retraining.
Recover-LoRA~\citep{das2025recoverlora} was originally developed for recovering accuracy from general weight corruption; here we demonstrate its effectiveness for structured quantization error, directly addressing the extension identified as future work in the original paper.

We evaluate accuracy recovery on Qwen3-4B across 12 benchmarks spanning commonsense reasoning, factual knowledge, and out-of-distribution tasks, extending Recover-LoRA~\citep{das2025recoverlora} with both synthetically generated data and a well-established finetuning dataset (OpenHermes). We additionally present throughput analysis across three model families (Qwen3-4B, Qwen3-14B, and GPT-OSS-20B) on two hardware platforms to establish the practical motivation for the W4/W2-GateUp configuration.
Our contributions are as follows:
\begin{enumerate}[leftmargin=*]
    \item We demonstrate that \textbf{Recover-LoRA effectively recovers accuracy from aggressive 2-bit quantization} in a Qwen3-4B case study using only 10k synthetic samples---without labeled data or full model retraining. Nine of twelve benchmarks (including out-of-distribution tasks) achieve 80--95\% accuracy recovery, with synthetic data performing comparably to curated labeled data. To demonstrate that the method is not specific to a single bit-width, we further validate Recover-LoRA at 3-bit gate/up precision in Appendix~\ref{app:3bit}, confirming positive recovery across bit-widths.

    \item We present a \textbf{practical end-to-end deployment pipeline} in which selective 2-bit quantization of gate and up projections delivers 7.5--23.3\% TPS gains over uniform W4 across three model families (4B--20B) and two hardware platforms, depending on context length, and show Recover-LoRA as the practical recovery mechanism for regaining the otherwise severely lost accuracy from 2-bit quantization.\end{enumerate}

\section{Related Work}
\label{sec:related}

\subsection{Weight Quantization for LLMs}

Post-training quantization (PTQ) methods compress LLM weights to lower precision without retraining.
AWQ~\citep{lin2024awq} identifies activation-aware salient weight channels and applies per-channel scaling before quantization, while GPTQ~\citep{frantar2023gptq} uses approximate second-order information to minimize layer-wise quantization error.
These methods are effective at 4-bit precision and can extend to 3-bit, but accuracy degrades substantially below 3 bits~\citep{zhu2024survey}.
Quantization-aware training (QAT) methods such as LLM-QAT~\citep{liu2023llmqat} and DL-QAT~\citep{ke2025dlqat} retrain the model with simulated quantization noise to improve robustness.
LLM-QAT notably demonstrates that synthetic data can replace labeled data for QAT via knowledge distillation.
However, true QAT requires simulating quantization during the forward pass and updating all model parameters, which is computationally expensive for large models and demands access to training infrastructure comparable to pretraining.
In the original Recover-LoRA work~\citep{das2025recoverlora}, a full-model finetuning baseline with synthetic data (denoted LLM-QAT*, not to be confused with true QAT) achieved negative accuracy recovery on degraded models, demonstrating that simply applying full-parameter distillation without quantization-aware gradient simulation can be counterproductive.
By contrast, Recover-LoRA's lightweight LoRA-based approach trains only a small number of adapter parameters on the quantized model as-is, avoiding the complexity of QAT while achieving positive recovery.
This makes it practical as a post-hoc deployment tool, as practitioners can apply standard PTQ methods and then recover accuracy without retraining or modifying the quantization pipeline.

Recent work on ultra-low-bit quantization includes QuIP\#~\citep{tseng2024quip} and AQLM~\citep{egiazarian2024aqlm}, which use incoherence processing and additive codebooks, respectively, to push quantization below 4 bits.
While these methods advance the quantization frontier, they require specialized kernels and may not integrate readily into existing deployment pipelines.
Recover-LoRA~\citep{das2025recoverlora} takes a complementary approach in that it focuses on recovering accuracy lost from PTQ, as opposed to improving the quantization algorithm itself.
In this work, we specifically extend Recover-LoRA to 2-bit quantization, demonstrating its effectiveness in recovering degraded accuracy from ultra-low-bit precision.

\subsection{Recover-LoRA}

Recover-LoRA~\citep{das2025recoverlora} was introduced as a data-free method to recover accuracy in functionally degraded language models.
The method trains LoRA adapters~\citep{hu2022lora} on selected model layers using logit distillation from the full-precision model, with synthetic data generated via hybrid sampling~\citep{liu2023llmqat}.
The original Recover-LoRA work demonstrated 5--17\% accuracy recovery on models degraded through improper weight serialization, outperforming both full-model distillation (LLM-QAT*) and supervised LoRA finetuning across multiple SLM architectures, and identified quantization-induced degradation as a natural next direction.
In this work, we apply Recover-LoRA to 2-bit quantized models, demonstrating its effectiveness beyond the synthetic error setting.

\subsection{Mixed-Precision Quantization}

Mixed-precision quantization assigns different bit-widths to different layers or components based on their sensitivity to quantization~\citep{dong2019hawq, wang2019haq}.
OWQ~\citep{lee2024owq} identifies outlier-vulnerable columns and assigns them higher precision.
Our W4/W2-GateUp strategy is a specific instance of mixed-precision quantization informed by throughput analysis. Gate and up projections are chosen for aggressive quantization because they constitute the majority of parameters in modern LLM architectures and are the primary bandwidth bottleneck during decode.
Notably, component-level sensitivity analyses show that MLP projections are in fact the most quantization-sensitive transformer components~\citep{cim2026diagnosing}, meaning that this throughput-optimal choice comes at a meaningful accuracy cost, further motivating the use of Recover-LoRA~\citep{das2025recoverlora} as a practical post-quantization accuracy recovery mechanism.

\subsection{Parameter-Efficient Fine-Tuning}

LoRA~\citep{hu2022lora} and its variants are widely used for parameter-efficient fine-tuning of LLMs.
QLoRA~\citep{dettmers2023qlora} enables finetuning of quantized models by backpropagating through a frozen 4-bit base model, while QA-LoRA~\citep{xu2024qalora} integrates quantization awareness into the LoRA training process.
However, these methods are designed for task adaptation with labeled data---they assume the quantized model is a starting point for downstream specialization, not that accuracy has been lost and needs to be restored.
Recover-LoRA uses LoRA in a fundamentally different setting, recovering the original model's general capabilities after quantization-induced degradation, using synthetic data and logit distillation rather than task-specific supervision.

\section{Background}
\label{sec:background}

\subsection{LoRA}

Low-Rank Adaptation (LoRA)~\citep{hu2022lora} augments a pretrained weight matrix $W \in \mathbb{R}^{d \times k}$ with two trainable low-rank matrices $A \in \mathbb{R}^{r \times k}$ and $B \in \mathbb{R}^{d \times r}$, where $r \ll \min(d, k)$.
The output of a LoRA-augmented layer is:
\begin{equation}
    Y = WX + \alpha BAX
    \label{eq:lora}
\end{equation}
where $X$ is the input activation and $\alpha$ is a scaling factor.
During training, $W$ is frozen and only $A$ and $B$ are updated.

\subsection{Knowledge Distillation}

Knowledge distillation~\citep{hinton2015distilling} trains a student model $M_S$ to match the output distribution of a teacher model $M_T$.
The training objective minimizes the KL divergence between teacher and student logit distributions:
\begin{equation}
    \mathcal{L}_{\text{KD}} = \text{KL}(p_t \| p_s) = \sum_i p_t^i \log \frac{p_t^i}{p_s^i}
    \label{eq:kd}
\end{equation}
where $p_t = \text{softmax}(M_T(x))$ and $p_s = \text{softmax}(M_S(x))$ for input $x$.

\subsection{W4/W2-GateUp Mixed-Precision Quantization}

Modern transformer architectures employ a gated MLP structure with gate, up, and down projection layers.
In models such as Qwen3~\citep{yang2025qwen3} and Llama~\citep{grattafiori2024llama3}, the gate and up projections typically account for the largest share of model parameters.
Our W4/W2-GateUp configuration quantizes the gate and up projection weights to 2-bit precision while keeping all other linear layers (attention projections, down projections, etc.) at 4-bit precision.
During decode, inference throughput is limited by memory bandwidth for loading model weights~\citep{patwari2025life}, so reducing weight precision on the layers with the largest parameter count yields the greatest throughput benefit.
We quantify this using LIFE~\citep{patwari2025life}, a roofline-based analytical framework that estimates decode TPS from the per-operator memory traffic across a full forward pass.
LIFE allows each layer to be configured independently, so W4/W2-GateUp is modeled by setting the gate and up projections to INT2 (group size 32) while all other linear layers use INT4 (group size 128).

However, this throughput-motivated selection comes with an accuracy trade-off.
Component-level sensitivity analyses across multiple model scales show that MLP up- and down-projection layers consistently exhibit the highest quantization sensitivity, while attention projections are substantially less sensitive~\citep{cim2026diagnosing}.
Aggressive 2-bit quantization of gate and up projections therefore incurs meaningful accuracy degradation.
We address this using Recover-LoRA~\citep{das2025recoverlora}, which trains lightweight LoRA adapters on the quantized layers via knowledge distillation from the full-precision model, as detailed in the following section.

\section{Recover-LoRA for 2-Bit Quantization}
\label{sec:method}

Our approach consists of two stages, selective mixed-precision quantization followed by accuracy recovery via Recover-LoRA.

\subsection{Stage 1: Selective Mixed-Precision Quantization}

Given a pretrained model $M_T$, we apply post-training quantization selectively to the gate and up projection layers.
Specifically, gate and up projections are quantized to INT2 with group size 32, while all other linear layers remain at full precision (BF16). A group size of 32 is chosen because finer grouping granularity reduces quantization error at low bit-widths by narrowing the dynamic range covered by each scale factor~\citep{dettmers2023case}.
This yields a BF16/W2-GateUp model $M_Q$, which we treat as our degraded student model $M_S$.
The selective quantization confines the accuracy loss to a known subset of layers, directly informing where to place LoRA adapters for recovery.
For practical deployment to edge devices, the remaining BF16 layers can be further quantized to INT4 using standard methods such as AWQ~\citep{lin2024awq}, which is widely demonstrated to be near-lossless at 4-bit precision, to realize the full W4/W2-GateUp throughput gains analyzed in Section~\ref{sec:throughput}.
We discuss the full deployment pipeline in Section~\ref{sec:deployment}.

\subsection{Stage 2: Accuracy Recovery via Recover-LoRA}

Following~\citet{das2025recoverlora}, we recover accuracy in $M_S$ by training LoRA adapters using logit distillation from the full-precision teacher $M_T$.
LoRA adapters are placed on the gate and up projection layers, the same layers that were quantized to 2-bit.
This placement targets the source of degradation directly, and empirically outperforms placing adapters on non-quantized layers (see Section~\ref{sec:experiments}).

Training data is generated using the hybrid sampling strategy from LLM-QAT~\citep{liu2023llmqat}, in which the pretrained model deterministically generates the first 3--5 tokens and then stochastically generates the remaining tokens up to a maximum sequence length.
This produces diverse, model-aligned training data without requiring any labeled datasets.
In Section~\ref{sec:experiments}, we also compare against training with curated labeled data to validate this choice.

The LoRA adapters $A$ and $B$ are optimized to minimize the KL divergence between the logit distributions of the quantized student $M_S$ (with adapters) and the full-precision teacher $M_T$:
\begin{equation}
    \min_{A, B} \; \mathbb{E}_{x \sim D_{\text{syn}}} \left[ \text{KL}\left(\text{softmax}(M_T(x)) \;\|\; \text{softmax}(M_S^{A,B}(x))\right) \right]
    \label{eq:recover}
\end{equation}
where $M_S^{A,B}$ denotes the quantized model augmented with LoRA adapters $A$ and $B$, and $D_{\text{syn}}$ is the synthetic training dataset.
During training, the quantized weights $W_S$ are frozen and only the LoRA parameters are updated.
After training, the adapters can be merged into the quantized weights or kept as a separate module during inference.

\subsection{Relation to Original Recover-LoRA}

The present work differs from the original Recover-LoRA~\citep{das2025recoverlora} in several ways.
Most fundamentally, the source of degradation is different. The original work used synthetic weight perturbation (improper serialization) to simulate functional degradation, whereas here the degradation comes from actual 2-bit post-training quantization.
Correspondingly, we place adapters on the quantized gate and up layers rather than on K/V or full attention+MLP layers as in the original work, since the degradation in this setting originates specifically from the 2-bit quantized projections.
Additionally, while the original Recover-LoRA used OLoRA initialization (SVD-based), we find that standard LoRA initialization yields more stable training for quantization error recovery, as OLoRA degrades when applied to quantized weights.
Finally, we demonstrate effectiveness with as few as 10k synthetic samples, compared to the 90--120k samples needed in the original work. While the two settings differ in several respects (model, adapter placement, initialization), this sample efficiency gap is consistent with the hypothesis that quantization-induced error may be more structured and thus easier to correct than random weight perturbation.

\section{Throughput Analysis}
\label{sec:throughput}

Before presenting accuracy recovery results, we establish the practical motivation for W4/W2-GateUp quantization through a comprehensive throughput analysis.
All throughput estimates are obtained via roofline modeling with LIFE~\citep{patwari2025life}.
For our experiments, LIFE is configured in \emph{fusion mode}~\citep[Sec.~3.2.1]{patwari2025life}, in which sequential operators within the MLP and attention blocks are modeled as fused kernels that keep intermediate tensors on-chip, matching the fused attention and fused gate-activation-down patterns used in modern serving stacks. Additionally, we leverage a modeled BF16 KV cache to isolate the benefits of TPS gains from weight quantization.
We sweep three model families, two hardware architectures, seven context lengths, and five quantization configurations.

\subsection{Models and Hardware}

We analyze three model families spanning different sizes and architectural profiles: Qwen3-4B~\citep{yang2025qwen3} (4B parameters), Qwen3-14B (14B parameters), and GPT-OSS-20B (20B parameters). The first two are dense transformers in which the gate and up projections constitute roughly 45--48\% of total parameters; GPT-OSS-20B is a Mixture-of-Experts model with a higher nominal gate/up fraction but only a subset of experts active per token, which limits realized throughput gains from quantizing those layers. We include an MoE model in our forecasting to characterize how selective 2-bit quantization behaves under MoE routing, where the gap between total and active parameters is substantial. Analysis is conducted on two architectures to isolate hardware sensitivity: Device~A (50~TOPs, 100~GB/s memory bandwidth) and Device~B (100~TOPs, 200~GB/s memory bandwidth).

\subsection{Throughput Results}
\label{sec:tps}

Tables~\ref{tab:tps_devicea} and~\ref{tab:tps_deviceb} summarize decode throughput across all three models on Device~A and Device~B, respectively.
We report two derived metrics: $\text{Gain\%} = (\text{TPS}_{\text{W4/W2}} - \text{TPS}_{\text{W4}}) / \text{TPS}_{\text{W4}}$, the relative TPS improvement of W4/W2-GateUp over the W4 baseline, and $\text{Capture\%} = (\text{TPS}_{\text{W4/W2}} - \text{TPS}_{\text{W4}}) / (\text{TPS}_{\text{W2}} - \text{TPS}_{\text{W4}})$, the fraction of the full uniform-W2 speedup that selective quantization captures.

\begin{table}[t]
\centering
\caption{Decode throughput (tokens per second) on Device~A (50~TOPs, 100~GB/s).}
\label{tab:tps_devicea}
\small
\begin{tabular}{lcccccc}
\toprule
\textbf{Model} & \textbf{Context} & \textbf{TPS W4} & \textbf{TPS W2} & \textbf{TPS W4/W2} & \textbf{Gain\%} & \textbf{Capture\%} \\
\midrule
\multirow{5}{*}{Qwen3-4B}
 & 1k   & 37.3 & 57.1 & 44.9 & 20.4\% & 38.5\% \\
 & 4k   & 32.8 & 47.1 & 38.5 & 17.5\% & 39.9\% \\
 & 8k   & 28.2 & 38.2 & 32.4 & 14.7\% & 41.4\% \\
 & 16k  & 22.1 & 27.8 & 24.5 & 11.2\% & 43.2\% \\
 & 32k  & 15.4 & 17.9 & 16.5 &  7.5\% & 45.1\% \\
\midrule
\multirow{5}{*}{Qwen3-14B}
 & 1k   & 10.3 & 16.4 & 12.7 & 23.0\% & 39.7\% \\
 & 4k   &  9.7 & 14.8 & 11.7 & 21.2\% & 40.4\% \\
 & 8k   &  8.9 & 13.1 & 10.6 & 19.2\% & 41.4\% \\
 & 16k  &  7.7 & 10.6 &  9.0 & 16.2\% & 43.1\% \\
 & 32k  &  6.1 &  7.7 &  6.8 & 12.3\% & 45.2\% \\
\midrule
\multirow{5}{*}{GPT-OSS-20B$^\dagger$}
 & 1k   & 27.9 & 50.6 & 31.8 & 14.1\% & 17.4\% \\
 & 4k   & 26.8 & 47.2 & 30.5 & 13.5\% & 17.8\% \\
 & 8k   & 25.5 & 43.3 & 28.8 & 12.8\% & 18.4\% \\
 & 16k  & 23.3 & 37.2 & 26.0 & 11.5\% & 19.3\% \\
 & 32k  & 19.8 & 29.0 & 21.7 &  9.6\% & 20.6\% \\
\bottomrule
\end{tabular}
\par\smallskip\noindent{\footnotesize $^\dagger$GPT-OSS-20B is a Mixture-of-Experts model; only a subset of experts are active per token, so the effective fraction of weights touched during inference is smaller than the total gate/up parameter count suggests. See text for discussion.}
\end{table}

\begin{table}[t]
\centering
\caption{Decode throughput (tokens per second) on Device~B (100~TOPs, 200~GB/s).}
\label{tab:tps_deviceb}
\small
\begin{tabular}{lcccccc}
\toprule
\textbf{Model} & \textbf{Context} & \textbf{TPS W4} & \textbf{TPS W2} & \textbf{TPS W4/W2} & \textbf{Gain\%} & \textbf{Capture\%} \\
\midrule
\multirow{5}{*}{Qwen3-4B}
 & 1k   & 56.3 & 85.7 & 67.7 & 20.1\% & 38.6\% \\
 & 4k   & 49.7 & 71.2 & 58.3 & 17.4\% & 40.0\% \\
 & 8k   & 42.9 & 58.1 & 49.2 & 14.7\% & 41.4\% \\
 & 16k  & 33.8 & 42.5 & 37.5 & 11.2\% & 43.2\% \\
 & 32k  & 23.7 & 27.6 & 25.5 &  7.6\% & 45.0\% \\
\midrule
\multirow{5}{*}{Qwen3-14B}
 & 1k   & 15.8 & 24.9 & 19.4 & 22.9\% & 39.7\% \\
 & 4k   & 14.8 & 22.5 & 17.9 & 21.2\% & 40.6\% \\
 & 8k   & 13.6 & 20.0 & 16.3 & 19.3\% & 41.5\% \\
 & 16k  & 11.8 & 16.3 & 13.8 & 16.3\% & 43.0\% \\
 & 32k  &  9.3 & 11.9 & 10.5 & 12.4\% & 44.8\% \\
\midrule
\multirow{5}{*}{GPT-OSS-20B$^\dagger$}
 & 1k   & 60.7 & 104.7 & 73.6 & 21.4\% & 29.5\% \\
 & 4k   & 57.5 &  95.7 & 69.1 & 20.1\% & 30.2\% \\
 & 8k   & 53.9 &  85.9 & 63.8 & 18.5\% & 31.1\% \\
 & 16k  & 47.7 &  71.3 & 55.4 & 16.1\% & 32.5\% \\
 & 32k  & 38.9 &  53.2 & 43.8 & 12.7\% & 34.5\% \\
\bottomrule
\end{tabular}
\par\smallskip\noindent{\footnotesize $^\dagger$GPT-OSS-20B is a Mixture-of-Experts model; see Table~\ref{tab:tps_devicea} footnote.}
\end{table}

\paragraph{Selective quantization captures a significant fraction of the full W2 speedup.}
As shown by the Capture\% column in Tables~\ref{tab:tps_devicea} and~\ref{tab:tps_deviceb}, the dense Qwen3 models consistently capture 38--45\% of the total TPS gain available from uniform W2 quantization, while quantizing only the gate and up projections. For GPT-OSS-20B the captured fraction is lower (17--21\% on Device~A, 29--35\% on Device~B) due to MoE routing effects discussed below. A substantial portion of the bandwidth benefit is therefore realized while confining quantization error to a well-defined subset of the model, making targeted accuracy recovery with Recover-LoRA feasible.

\paragraph{Dense models gain 20--23\% TPS over W4 at short contexts, tapering with length.}
For the dense Qwen3 models, W4/W2-GateUp delivers 20--23\% TPS improvement at short context lengths (1k tokens), tapering to 7--12\% at 32k tokens.
These gains are moderated by non-weight operators (KV cache reads, RMSNorm, rotary embeddings, attention BMMs, softmax) whose bandwidth cost is fixed regardless of weight precision, and by the finer group size used for INT2 (32 versus 128 for INT4), which increases per-weight metadata overhead.

\paragraph{MoE routing and group-size effects compress gains for GPT-OSS-20B.}
GPT-OSS-20B achieves more modest W4/W2-GateUp gains on Device~A (14.1\% at 1k, declining to 9.6\% at 32k) because its MoE routing activates only a subset of experts per token, so the effective weight bandwidth during inference is much smaller than the total parameter count suggests.
GPT-OSS-20B also uses a smaller INT4 group size of 64 (versus 128 for the dense models), which further narrows the metadata gap between INT4 and INT2 and compresses the relative gain.

\paragraph{KV cache bandwidth dominates at long contexts, diminishing weight-quantization gains.}
Gains diminish at longer context lengths across all models.
As context length increases, the BF16 KV cache grows to dominate memory bandwidth consumption, reducing the relative impact of weight compression on overall throughput.
For example, Qwen3-4B's W4/W2-GateUp gain drops from 20.4\% at 1k to 7.5\% at 32k.
This behavior suggests that combining W4/W2-GateUp with KV cache quantization could yield further improvements at long contexts, which we leave to future work.

\paragraph{Relative TPS gains for dense models are consistent across hardware architectures.}
Comparing Tables~\ref{tab:tps_devicea} and~\ref{tab:tps_deviceb}, Device~B exhibits uniformly higher absolute TPS values, as expected from its greater memory bandwidth, but Gain\% and Capture\% remain nearly identical for the dense models (within $\pm$0.5 percentage points). GPT-OSS-20B shows notably higher Gain\% and Capture\% on Device~B (e.g., 21.4\% Gain and 29.5\% Capture at 1k versus 14.1\% and 17.4\% on Device~A), indicating that the MoE throughput profile is more hardware-sensitive. Beyond peak TOPs and bandwidth, the two devices differ in compute and memory utilization efficiency, which shifts the compute-to-memory-bound transition and disproportionately affects MoE models where active weight footprint per token is smaller. Full seven-context-length results for both devices are provided in Appendix~\ref{app:throughput}.

\section{Accuracy Recovery Experiments}
\label{sec:experiments}

We present a case study on Qwen3-4B~\citep{yang2025qwen3} to demonstrate the effectiveness of Recover-LoRA for recovering accuracy from 2-bit quantization.

\subsection{Experimental Setup}

Our teacher model $M_T$ is the pretrained Qwen3-4B in full precision (BF16).
The student model $M_S$ is the Qwen3-4B with gate and up projections quantized to INT2 (group size 32, chosen to reduce quantization error at low bit-widths as discussed in Section~\ref{sec:method}) while all remaining layers are kept at BF16.
This configuration isolates the effect of 2-bit quantization on the gate/up layers, allowing us to evaluate Recover-LoRA's ability to recover from this specific source of degradation without confounding effects from quantizing other layers.

\subsubsection{Hyperparameters}
Table~\ref{tab:hyperparams} summarizes the training configuration.
We train LoRA adapters on the gate and up projection layers with a rank of 32, alpha of 64, learning rate of 1e-4, and standard LoRA initialization.
The training uses 10k samples and optimizes via KL divergence against the teacher model's logit distribution.
We study two training data conditions separately.
The first uses curated labeled data from OpenHermes, a widely used finetuning dataset covering a broad distribution of general knowledge tasks.
The second uses synthetically generated data via the hybrid sampling strategy from Recover-LoRA~\citep{das2025recoverlora} and LLM-QAT~\citep{liu2023llmqat}, in which the full-precision model deterministically generates the first 3--5 tokens of each sample and then stochastically generates the remaining tokens up to a maximum sequence length, producing diverse, model-aligned data without any labeled source.
We compare both conditions to evaluate whether Recover-LoRA can achieve comparable recovery without access to labeled datasets.

\begin{table}[t]
\centering
\caption{Recover-LoRA training hyperparameters for the Qwen3-4B case study.}
\label{tab:hyperparams}
\small
\begin{tabular}{ll}
\toprule
\textbf{Hyperparameter} & \textbf{Value} \\
\midrule
Training samples        & 10,000 \\
LoRA target layers      & Gate \& Up projections \\
Learning rate           & 1e-4 \\
LoRA rank ($r$)         & 32 \\
LoRA alpha ($\alpha$)   & 64 \\
LoRA initialization     & Standard (Kaiming uniform for $A$, zero for $B$)\footnote{\url{https://huggingface.co/docs/peft/developer_guides/lora}} \\
Training loss           & KL divergence (logit distillation) \\
Training data           & Synthetic (hybrid sampling) \\
\bottomrule
\end{tabular}
\end{table}

\subsubsection{Evaluation Datasets}
We evaluate on nine in-distribution benchmarks spanning commonsense reasoning and factual knowledge, including HellaSwag~\citep{zellers2019hellaswag}, MMLU~\citep{hendrycks2021mmlu}, TinyMMLU, ARC Challenge~\citep{clark2018arc}, WinoGrande~\citep{sakaguchi2021winogrande}, PiQA~\citep{bisk2020piqa}, OpenBookQA~\citep{mihaylov2018openbookqa}, BoolQ~\citep{clark2019boolq}, and TinyGSM8k.
These tasks are considered in-distribution in the sense that their domains (commonsense reasoning, factual knowledge, reading comprehension) overlap with the types of content represented in the OpenHermes training data.
We additionally evaluate on three out-of-distribution (OOD) benchmarks, LAMBADA, TruthfulQA, and LogiQA2, whose task formats (cloze completion, truthfulness evaluation, logical reasoning) are not directly represented in either training set, to assess whether recovery generalizes beyond the training data distribution (see Section~\ref{sec:ood}).
The in-distribution versus OOD distinction is most relevant for the OpenHermes-trained model; for the synthetically trained model, no task-specific data is used, so in principle all benchmarks are out-of-distribution. Nevertheless, we report Recover-LoRA accuracy recovery across all 12 tasks for completeness.
\subsubsection{Accuracy Recovery Metric}
We measure accuracy recovery using the Accuracy Recovery Percentage (AR\%)~\citep{das2025recoverlora}, defined as:
\begin{equation}
    \text{AR\%} = \frac{E_S^* - E_S}{|E_S - E_T|} \times 100
\end{equation}
where $E_T$, $E_S$, and $E_S^*$ are evaluation scores of the teacher, quantized student, and recovered student, respectively.
An AR\% of 100 indicates full recovery to teacher-level accuracy, while an AR\% of 0 indicates no improvement over the quantized model.

\subsection{Accuracy Recovery Results}

Table~\ref{tab:recovery} presents accuracy recovery results on Qwen3-4B with two training data sources, a curated labeled dataset (OpenHermes-10k) and synthetic data (Syn-10k).
The results below focus on 2-bit gate/up quantization as the more aggressive and challenging setting; we additionally report recovery at 3-bit precision in Appendix~\ref{app:3bit}.

\begin{table}[t]
\centering
\caption{Accuracy recovery on Qwen3-4B with W2-GateUp quantization (gate/up at INT2, remaining layers at BF16). $E_T$: teacher (BF16) accuracy, $E_S$: quantized student accuracy, $E_S^*$: recovered student accuracy, AR\%: accuracy recovery percentage. Both recovery configurations use 10k training samples with identical LoRA hyperparameters; only the training data source differs.}
\label{tab:recovery}
\small
\begin{tabular}{lcc|cc|cc}
\toprule
\textbf{Eval Task} & \textbf{$E_T$} & \textbf{$E_S$} & \multicolumn{2}{c|}{\textbf{OpenHermes-10k}} & \multicolumn{2}{c}{\textbf{Synthetic-10k}} \\
 & & & \textbf{$E_S^*$} & \textbf{AR\%} & \textbf{$E_S^*$} & \textbf{AR\%} \\
\midrule
HellaSwag       & 52.2 & 28.0 & 48.3 & 83.9 & 47.4 & 80.3 \\
MMLU            & 68.3 & 24.5 & 61.2 & 83.7 & 60.5 & 82.2 \\
TinyMMLU        & 69.2 & 26.9 & 66.1 & 92.7 & 60.5 & 79.4 \\
ARC Challenge   & 50.4 & 19.0 & 43.2 & 76.9 & 45.0 & 82.6 \\
WinoGrande      & 66.3 & 50.7 & 64.0 & 85.3 & 64.5 & 88.4 \\
PiQA            & 75.0 & 54.3 & 72.6 & 88.2 & 72.6 & 88.4 \\
OpenBookQA      & 29.6 & 12.4 & 27.8 & 89.5 & 28.0 & 90.7 \\
BoolQ           & 84.9 & 38.4 & 82.0 & 93.8 & 82.5 & 94.9 \\
TinyGSM8k       & 56.3 &  1.5 & 31.7 & 55.0 & 27.8 & 48.0 \\
\bottomrule
\end{tabular}
\end{table}

\paragraph{2-bit quantization causes severe degradation, but Recover-LoRA achieves strong recovery.}
The raw accuracy numbers reveal the severity of 2-bit gate/up quantization. Some tasks are severely degraded, with MMLU falling from 68.3\% to 24.5\% and TinyGSM8k from 56.3\% to 1.5\%.
Despite this, Recover-LoRA achieves strong recovery using only 10k synthetic samples.
Seven of the nine in-distribution benchmarks exceed 80\% AR\%, and BoolQ and OpenBookQA both exceed 90\%. Including the out-of-distribution results (Section~\ref{sec:ood}), 9 of 12 total benchmarks achieve 80--95\% recovery.
This level of recovery with only 10k samples is notable given that the original Recover-LoRA required 90--120k samples for positive recovery on synthetically degraded models, suggesting that quantization-induced error may be more structured and thus easier to correct than random weight perturbation.

\paragraph{Synthetic data performs comparably to curated labeled data.}
Across the nine in-distribution benchmarks, the curated OpenHermes-10k and synthetic datasets achieve comparable AR\%, with neither consistently dominating the other.
On five of the nine benchmarks (ARC Challenge, WinoGrande, PiQA, OpenBookQA, and BoolQ), synthetic data matches or exceeds the labeled dataset in AR\%.
The practical implication is that practitioners can achieve strong accuracy recovery without access to any curated datasets, relying entirely on the model itself to generate training data. We examine why this holds in the out-of-distribution analysis below.

\paragraph{Mathematical reasoning remains challenging to recover.}
TinyGSM8k shows the lowest AR\% at 48--55\%. Mathematical reasoning appears particularly susceptible to degradation under 2-bit quantization, with TinyGSM8k accuracy dropping from 56.3\% to 1.5\%, and may benefit from task-specific complementary recovery techniques. Nevertheless, Recover-LoRA brings this back to 27.8--31.7\%, a substantial absolute improvement even though the AR\% appears modest relative to other benchmarks.

\paragraph{Recovery generalizes to out-of-distribution tasks.}
\label{sec:ood}
Table~\ref{tab:ood} presents results on three OOD evaluation benchmarks, LAMBADA, TruthfulQA, and LogiQA2.

\begin{table}[t]
\centering
\caption{Out-of-distribution accuracy recovery on Qwen3-4B with W2-GateUp quantization (gate/up at INT2, remaining layers at BF16). $E_T$: teacher (BF16) accuracy, $E_S$: quantized student accuracy, $E_S^*$: recovered student accuracy. Evaluation tasks were not represented in or similar to the training data.}
\label{tab:ood}
\small
\begin{tabular}{lcc|cc|cc}
\toprule
\textbf{Eval Task} & \textbf{$E_T$} & \textbf{$E_S$} & \multicolumn{2}{c|}{\textbf{OpenHermes-10k}} & \multicolumn{2}{c}{\textbf{Synthetic-10k}} \\
 & & & \textbf{$E_S^*$} & \textbf{AR\%} & \textbf{$E_S^*$} & \textbf{AR\%} \\
\midrule
LAMBADA (OpenAI)  & 59.3 &  3.4 & 55.6 & 93.3 & 53.4 & 89.5 \\
TruthfulQA        & 54.8 & 48.7 & 53.3 & 75.6 & 54.2 & 90.7 \\
LogiQA2           & 34.4 & 24.7 & 32.9 & 84.9 & 30.5 & 60.5 \\
\bottomrule
\end{tabular}
\end{table}

These tasks are out-of-distribution with respect to the OpenHermes training set, meaning the OpenHermes-trained model has not seen such task formats or domains during recovery training.

LAMBADA exhibits the most severe degradation among the OOD tasks, with accuracy falling from 59.3\% to 3.4\%, yet Recover-LoRA recovers it to 53.4--55.6\%, achieving over 89\% AR\% under both training data conditions.
Overall, OOD recovery is comparable to in-distribution performance under both data conditions, indicating that Recover-LoRA is recovering general model capabilities through alignment with the teacher's output distribution rather than memorizing patterns from the training data.
This is consistent with the distillation-based nature of the method, where the training objective aligns the student's logit distribution with the teacher's across a broad range of inputs, and the resulting adapters generalize beyond the specific content of the training data.

\subsection{Practical Training Considerations}

Through our experiments, we identify several practical considerations for applying Recover-LoRA to quantized models.

The original Recover-LoRA used OLoRA initialization, which performs SVD on the original pretrained weights to initialize the LoRA adapters.
While this works well when the underlying weights are clean, we find that OLoRA degrades when applied to quantized (QDQ) weights and can diverge at higher learning rates.
We instead use standard LoRA initialization (Kaiming uniform for $A$, zero for $B$), which provides more stable training and faster convergence for quantization recovery.
We attribute this to the fact that OLoRA's SVD is computed on the corrupted quantized weights, producing a suboptimal initialization that biases early training, whereas standard initialization makes no assumptions about the weight structure and allows the optimizer to learn the correction from a neutral starting point.

With respect to adapter placement, we observe that placing adapters on the quantized (QDQ) layers consistently outperforms placing them on non-quantized layers.
This result is intuitive in that the adapters directly compensate for the quantization error at its source. It also has a practical benefit, as the adapter placement decision is straightforward and does not require expensive search since the quantized layers are known a priori.

Finally, we note that each quantized model requires its own trained LoRA adapter, as the quantization error pattern is configuration-specific. However, because Recover-LoRA trains only low-rank adapters on a small synthetic dataset, the per-configuration training cost is modest relative to full model retraining or quantization-aware training.

\section{Deployment Pipeline}
\label{sec:deployment}

We envision Recover-LoRA as a standard post-quantization step in the model deployment pipeline:
\begin{enumerate}[leftmargin=*]
    \item \textbf{Quantize} the gate and up projections to INT2.
    \item \textbf{Generate} 10k synthetic samples from the full-precision model via hybrid sampling.
    \item \textbf{Train} LoRA adapters on the quantized gate/up layers using logit distillation (single-GPU, low cost).
    \item \textbf{Quantize} remaining layers to INT4 for the full W4/W2-GateUp deployment configuration.
    \item \textbf{Deploy} the quantized model with merged or attached LoRA adapters.
\end{enumerate}
This pipeline requires access to the full-precision model as a teacher during the knowledge distillation LoRA finetuning, but no labeled data, making it applicable in proprietary or data-scarce settings.
The degree of accuracy recovery is tunable at Steps~2 and~3 through synthetic data volume and composition (Step~2) and hyperparameter selection such as rank, learning rate, and training duration (Step~3), offering practitioners a flexible knob for trading off training cost against accuracy.

Note that our accuracy recovery experiments (Section~\ref{sec:experiments}) use a BF16/W2-GateUp configuration (gate/up at INT2, remaining layers at BF16) to isolate the effect of 2-bit quantization on the gate/up layers without confounding effects from quantizing other layers.
In practice, deploying on edge devices with limited memory requires quantizing the remaining layers as well. Step 4 applies INT4 quantization to these layers using standard PTQ methods such as AWQ~\citep{lin2024awq}, which has been widely shown to be near-lossless at 4-bit precision. More recent INT4 methods such as ParoQuant~\citep{liang2026paroquant}, which has demonstrated even higher accuracy than AWQ on reasoning tasks, represent promising alternatives as the base quantizer for this step.

\section{Conclusion}
We have demonstrated that Recover-LoRA~\citep{das2025recoverlora}, a lightweight, data-free accuracy recovery method, can serve as a practical post-quantization recovery tool for aggressive 2-bit weight quantization.
Our selective W4/W2-GateUp quantization strategy targets gate and up projections for 2-bit quantization, deliberately quantizing the layers that yield the largest throughput gains despite their known sensitivity to low-bit compression.
Across the three model families studied, this achieves 7.5--23.3\% TPS improvement over uniform 4-bit quantization depending on model and context length, while confining accuracy degradation to a predictable subset of layers.
Recover-LoRA then recovers a substantial portion of the lost accuracy using only 10k synthetic training samples, achieving 80--95\% recovery on 9 of 12 benchmarks including out-of-distribution tasks.
Synthetic data performs comparably to curated labeled data for distillation-based recovery, reducing the dependency on high-quality datasets.

Together, these results establish Recover-LoRA as a viable post-quantization recovery mechanism for aggressive sub-4-bit compression, validated at both 2-bit and 3-bit (Appendix~\ref{app:3bit}) precision. Recovery is further tunable through hyperparameter selection, data scale, and data composition (Section~\ref{sec:deployment}), offering practitioners a practical knob for balancing training cost against recovered accuracy.

\subsection{Limitations and Future Work}
\label{sec:limitations}
Our accuracy recovery experiments focus on a single model (Qwen3-4B) with BF16 for non-gate/up layers. The most immediate next step is validating recovery under the full W4/W2-GateUp deployment configuration, including alternative INT4 base quantizers such as ParoQuant~\citep{liang2026paroquant}. Additionally, extending the accuracy recovery evaluation to additional model families and scales remains important for establishing robustness. Our results also showed that mathematical reasoning recovery lags behind other benchmarks and may benefit from domain-targeted finetuning data or complementary recovery techniques. Therefore, additional analyses on increased data scale to measure improvement on harder tasks is another important direction. Combining W4/W2-GateUp with KV cache quantization could yield further throughput improvements at longer context lengths, provided recovery quality is maintained.
\bibliography{references}

\clearpage
\appendix

\section{Detailed Throughput Results}
\label{app:throughput}

Tables~\ref{tab:tps_full_devicea} and~\ref{tab:tps_full_deviceb} provide the complete throughput results on Device~A and Device~B, respectively, across all three models and seven context lengths.
These results confirm that the trends observed in the main text hold consistently across both hardware platforms.
Device~B exhibits uniformly higher absolute TPS values but similar relative gain percentages, indicating that the W4/W2-GateUp strategy is hardware-agnostic in its relative benefits.
$^\dagger$GPT-OSS-20B is a Mixture-of-Experts model; see Section~\ref{sec:tps} for discussion of its gain profile.

\begin{table}[h]
\centering
\caption{Complete TPS results on Device~A (50~TOPs, 100~GB/s) across all models and context lengths.}
\label{tab:tps_full_devicea}
\small
\begin{tabular}{lccccr}
\toprule
\textbf{Model} & \textbf{Context} & \textbf{W4} & \textbf{W2} & \textbf{W4/W2} & \textbf{Gain\%} \\
\midrule
\multirow{7}{*}{Qwen3-4B}
 & 512  & 38.2 & 59.2 & 46.2 & 21.0\% \\
 & 1k   & 37.3 & 57.1 & 44.9 & 20.4\% \\
 & 2k   & 35.7 & 53.3 & 42.5 & 19.3\% \\
 & 4k   & 32.8 & 47.1 & 38.5 & 17.5\% \\
 & 8k   & 28.2 & 38.2 & 32.4 & 14.7\% \\
 & 16k  & 22.1 & 27.8 & 24.5 & 11.2\% \\
 & 32k  & 15.4 & 17.9 & 16.5 &  7.5\% \\
\midrule
\multirow{7}{*}{Qwen3-14B}
 & 512  & 10.5 & 16.6 & 12.9 & 23.3\% \\
 & 1k   & 10.3 & 16.4 & 12.7 & 23.0\% \\
 & 2k   & 10.1 & 15.8 & 12.4 & 22.4\% \\
 & 4k   &  9.7 & 14.8 & 11.7 & 21.2\% \\
 & 8k   &  8.9 & 13.1 & 10.6 & 19.2\% \\
 & 16k  &  7.7 & 10.6 &  9.0 & 16.2\% \\
 & 32k  &  6.1 &  7.7 &  6.8 & 12.3\% \\
\midrule
\multirow{7}{*}{GPT-OSS-20B$^\dagger$}
 & 512  & 28.1 & 51.2 & 32.1 & 14.2\% \\
 & 1k   & 27.9 & 50.6 & 31.8 & 14.1\% \\
 & 2k   & 27.5 & 49.4 & 31.4 & 13.9\% \\
 & 4k   & 26.8 & 47.2 & 30.5 & 13.5\% \\
 & 8k   & 25.5 & 43.3 & 28.8 & 12.8\% \\
 & 16k  & 23.3 & 37.2 & 26.0 & 11.5\% \\
 & 32k  & 19.8 & 29.0 & 21.7 &  9.6\% \\
\bottomrule
\end{tabular}
\end{table}

\begin{table}[h]
\centering
\caption{Complete TPS results on Device~B (100~TOPs, 200~GB/s) across all models and context lengths.}
\label{tab:tps_full_deviceb}
\small
\begin{tabular}{lccccr}
\toprule
\textbf{Model} & \textbf{Context} & \textbf{W4} & \textbf{W2} & \textbf{W4/W2} & \textbf{Gain\%} \\
\midrule
\multirow{7}{*}{Qwen3-4B}
 & 512  & 57.6 & 88.7 & 69.5 & 20.7\% \\
 & 1k   & 56.3 & 85.7 & 67.7 & 20.1\% \\
 & 2k   & 53.9 & 80.3 & 64.2 & 19.1\% \\
 & 4k   & 49.7 & 71.2 & 58.3 & 17.4\% \\
 & 8k   & 42.9 & 58.1 & 49.2 & 14.7\% \\
 & 16k  & 33.8 & 42.5 & 37.5 & 11.2\% \\
 & 32k  & 23.7 & 27.6 & 25.5 &  7.6\% \\
\midrule
\multirow{7}{*}{Qwen3-14B}
 & 512  & 15.9 & 25.3 & 19.6 & 23.2\% \\
 & 1k   & 15.8 & 24.9 & 19.4 & 22.9\% \\
 & 2k   & 15.4 & 24.0 & 18.9 & 22.3\% \\
 & 4k   & 14.8 & 22.5 & 17.9 & 21.2\% \\
 & 8k   & 13.6 & 20.0 & 16.3 & 19.3\% \\
 & 16k  & 11.8 & 16.3 & 13.8 & 16.3\% \\
 & 32k  &  9.3 & 11.9 & 10.5 & 12.4\% \\
\midrule
\multirow{7}{*}{GPT-OSS-20B$^\dagger$}
 & 512  & 61.2 & 106.3 & 74.5 & 21.6\% \\
 & 1k   & 60.7 & 104.7 & 73.6 & 21.4\% \\
 & 2k   & 59.6 & 101.5 & 72.1 & 20.9\% \\
 & 4k   & 57.5 &  95.7 & 69.1 & 20.1\% \\
 & 8k   & 53.9 &  85.9 & 63.8 & 18.5\% \\
 & 16k  & 47.7 &  71.3 & 55.4 & 16.1\% \\
 & 32k  & 38.9 &  53.2 & 43.8 & 12.7\% \\
\bottomrule
\end{tabular}
\end{table}

\section{Accuracy Recovery at 3-Bit Precision}
\label{app:3bit}

To validate that Recover-LoRA generalizes beyond the 2-bit setting examined in the main text, we repeat the accuracy recovery experiment on Qwen3-4B with gate and up projections quantized to 3-bit (INT3, group size 32) while all other layers remain at BF16.
Table~\ref{tab:recovery_3bit} reports the results.

\begin{table}[h]
\centering
\caption{Accuracy recovery on Qwen3-4B with 3-bit gate/up quantization (BF16/W3-GateUp). $E_T$ = BF16 teacher, $E_S$ = W3 quantized student, $E_S^*$ = recovered. AR\% = accuracy recovery percentage.}
\label{tab:recovery_3bit}
\small
\begin{tabular}{lcccccc}
\toprule
\textbf{Benchmark} & $E_T$ & $E_S$ & \multicolumn{2}{c}{\textbf{OpenHermes-10k}} & \multicolumn{2}{c}{\textbf{Synthetic-10k}} \\
\cmidrule(lr){4-5} \cmidrule(lr){6-7}
 & & & $E_S^*$ & AR\% & $E_S^*$ & AR\% \\
\midrule
HellaSwag    & 52.2 & 49.6 & 51.3 & 65.3 & 50.9 & 52.3 \\
MMLU         & 68.3 & 65.6 & 66.5 & 35.2 & 66.9 & 47.6 \\
TinyMMLU     & 69.2 & 62.2 & 67.0 & 68.2 & 62.5 &  4.3 \\
ARC-C        & 50.4 & 46.5 & 48.6 & 54.2 & 48.8 & 58.8 \\
WinoGrande   & 66.3 & 64.2 & 66.9 & 129.6 & 68.0 & 181.2 \\
PiQA         & 75.0 & 72.5 & 75.0 & 97.6 & 74.1 & 62.9 \\
OpenbookQA   & 29.6 & 28.8 & 30.6 & 225.0 & 29.4 & 75.0 \\
BoolQ        & 84.9 & 83.1 & 84.1 & 57.8 & 85.0 & 105.0 \\
TinyGSM8k    & 56.3 & 39.4 & 42.6 & 19.1 & 47.7 & 49.1 \\
\midrule
\multicolumn{7}{l}{\emph{Out-of-distribution}} \\
\midrule
lambada\_openai & 59.3 & 54.7 & 58.9 & 91.7 & 58.7 & 87.4 \\
truthfulqa$^\dagger$ & 54.8 & 55.6 & 55.7 & --- & 55.0 & --- \\
logiqa2         & 34.4 & 32.4 & 33.7 & 64.5 & 32.8 & 22.3 \\
\bottomrule
\end{tabular}
\vspace{2pt}
\par\noindent{\footnotesize $^\dagger$TruthfulQA's quantized accuracy exceeds the teacher (55.6 versus 54.8), so the teacher-student gap is inverted and AR\% is undefined.}
\end{table}

Recovery is positive across the majority of benchmarks, confirming that Recover-LoRA is effective at 3-bit precision as well.
TruthfulQA's quantized accuracy slightly exceeds the teacher (55.6 versus 54.8), indicating negligible degradation on that task at 3-bit; AR\% is therefore undefined for this benchmark.
Several benchmarks with small teacher-student gaps (e.g., WinoGrande, OpenbookQA, BoolQ) show AR\% exceeding 100\%, indicating that the recovered model slightly surpasses the teacher on those tasks, likely due to beneficial regularization from the LoRA adapters.

At 3-bit, the quantization degradation is milder than at 2-bit, so the absolute error signal available to the LoRA adapters during distillation is smaller. This appears to widen the gap between OpenHermes and Synthetic recovery, in contrast to the 2-bit setting where the two data sources achieved more similar results.
The in-distribution versus out-of-distribution distinction is most relevant for the OpenHermes-trained model; for the synthetically trained model, no task-specific data is used, so in principle all benchmarks are out-of-distribution.
LogiQA2 shows weaker recovery with synthetic data (22.3\% AR\%) compared to OpenHermes (64.5\%), suggesting that logical reasoning tasks may require more targeted or larger-scale synthetic data for effective recovery at milder quantization levels.

\end{document}